\pdfoutput=1

\documentclass[11pt]{article}

\usepackage{EACL2023}

\usepackage{times}
\usepackage{latexsym}

\usepackage[T1]{fontenc}

\usepackage[utf8]{inputenc}

\usepackage{microtype}

\usepackage{inconsolata}

\usepackage{booktabs}
\usepackage{multirow}

\usepackage[pdftex]{graphicx}
\usepackage{listings}

\usepackage{tikz}
\def\checkmark{\tikz\fill[scale=0.4](0,.35) -- (.25,0) -- (1,.7) -- (.25,.15) -- cycle;} 

\definecolor{codegreen}{rgb}{0,0.6,0}
\definecolor{codegray}{rgb}{0.5,0.5,0.5}
\definecolor{codepurple}{rgb}{0.58,0,0.82}
\definecolor{backcolour}{rgb}{0.95,0.95,0.92}

\lstdefinestyle{mystyle}{
    backgroundcolor=\color{backcolour},   
    commentstyle=\color{codegreen},
    keywordstyle=\color{magenta},
    numberstyle=\tiny\color{codegray},
    stringstyle=\color{codepurple},
    basicstyle=\ttfamily\footnotesize,
    breakatwhitespace=false,         
    breaklines=true,                 
    captionpos=b,                    
    keepspaces=true,                 
    numbers=left,                    
    numbersep=5pt,                  
    showspaces=false,                
    showstringspaces=false,
    showtabs=false,                  
    tabsize=2
}

\title{Machine Translation between Spoken Languages and Signed Languages Represented in SignWriting}

\author{Zifan Jiang \\
  University of Zurich \\
  \texttt{jiang@cl.uzh.ch} \\
  \And
  Amit Moryossef \\
  Bar-Ilan University \\
  University of Zurich \\
  \texttt{amitmoryossef@gmail.com} \\
  \And
  Mathias Müller \\
  University of Zurich \\
  \texttt{mmueller@cl.uzh.ch}
  \And
  Sarah Ebling \\
  University of Zurich \\
  \texttt{ebling@cl.uzh.ch}
}

\begin{document}
\maketitle
\begin{abstract}
This paper presents work on novel machine translation (MT) systems between spoken and signed languages, where signed languages are represented in SignWriting, a sign language writing system. Our work\footnote{Code and documentation available at \url{https://github.com/J22Melody/signwriting-translation}} seeks to address the lack of out-of-the-box support for signed languages in current MT systems and is based on the SignBank dataset, which contains pairs of spoken language text and SignWriting content. We introduce novel methods to parse, factorize, decode, and evaluate SignWriting, leveraging ideas from neural factored MT. In a bilingual setup---translating from American Sign Language to (American) English---our method achieves over 30 BLEU, while in two multilingual setups--- translating in both directions between spoken languages and signed languages---we achieve over 20 BLEU. We find that common MT techniques used to improve spoken language translation similarly affect the performance of sign language translation. These findings validate our use of an intermediate text representation for signed languages to include them in NLP research.
\end{abstract}

\section{Introduction}

Most current machine translation (MT) systems only support spoken language input and output (text or speech), which excludes around 200 different signed languages used by up to 70 million deaf people\footnote{According to the World Federation of the Deaf:
\url{https://wfdeaf.org/our-work/}} worldwide from modern language technology.
Since signed languages are also natural languages, \citet{yin-etal-2021-including} calls for including sign language processing (SLP) in natural language processing (NLP) research.

From a technical point of view, SLP brings novel challenges to NLP due to the visual-gestural modality of sign language and special linguistic features (e.g., the use of space, simultaneity, referencing), which requires both computer vision (CV) and NLP technologies. Crucially, the lack of a standardized or widely used written form for signed languages has hindered their inclusion in NLP research.

\begin{figure}
    \centering
    \includegraphics[width=\linewidth]{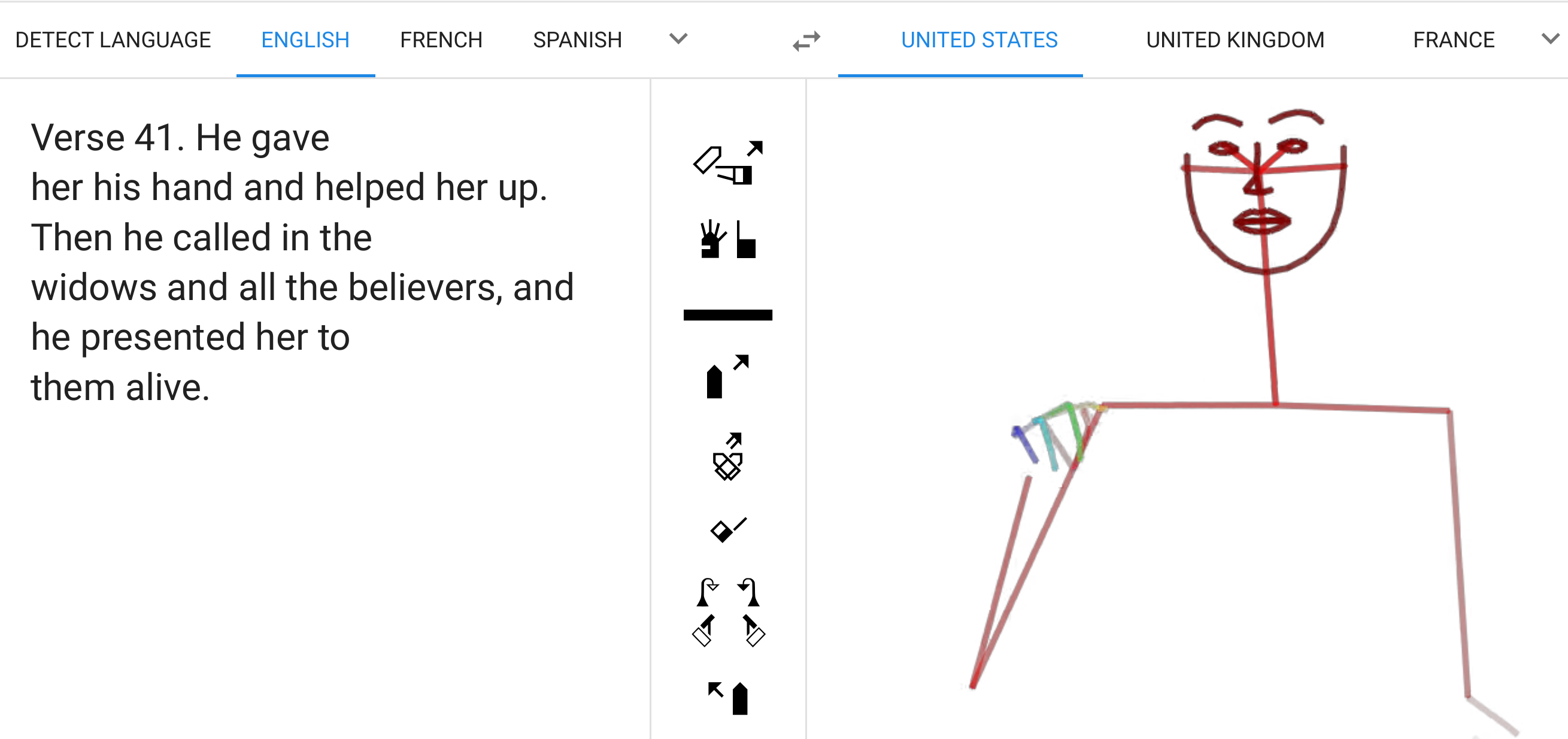}
    \caption{Demo application based on our models, translating from spoken languages to signed languages represented in SignWriting, then to human poses.}
    \label{fig:demo}
\end{figure}

However, sign language writing systems do exist and are sporadically used (e.g., SignWriting \citep{writing:sutton1990lessons} and HamNoSys  \citep{writing:prillwitz1990hamburg}). Therefore, we adopt the proposal of \citet{yin-etal-2021-including} to formulate the sign language translation (SLT) task using a sign language writing system as an intermediate step (illustrated by Figure \ref{fig:demo}): given spoken language text, we propose to translate to sign language in a written form, then transform this intermediate result into a final video or pose output\footnote{Note that the second step, animation of SignWriting into human poses or video, is not included in this research. In the demo application, spoken language text is translated directly into sign language poses, resulting in low-quality output.}---and vice versa. According to this multi-step view of SLT, in this work we study translation between signed languages in written form and spoken languages. We use \textit{SignWriting} as the intermediate writing system.

SignWriting has many advantages, like being universal (multilingual), comparatively easy to understand, extensively documented, and computer-supported. In addition, despite looking pictographic, it is a well-defined writing system. Every sign can be written as a sequence of symbols (box markers, graphemes, and punctuation marks) and their location on a 2-dimensional plane.

To our knowledge, this work is the first to create automatic SLT systems that use SignWriting. 
Our main contributions are as follows: (a) we propose methods to parse (\S\ref{sec:parse_fsw}), factorize (\S\ref{sec:factored}), decode (\S\ref{subsec:model_3}), and evaluate (\S\ref{subsec:model_3}) SignWriting sequences;
(b) we report experiments on multilingual machine translation systems between SignWriting and spoken language text (\S\ref{sec:exp});
(c) we demonstrate that common techniques for low-resource MT are beneficial for SignWriting translation systems (\S\ref{sec:discussion}).



\section{Background}

\subsection{Sign language processing (SLP)}

SLP \citep{bragg2019sign,yin-etal-2021-including,moryossef2021slp} is an emerging subfield of both NLP and CV, which focuses on automatic processing and analysis of sign language content. Prominent tasks include pose estimation from sign language videos \citep{pose:cao2017realtime, pose:cao2018openpose, pose:alp2018densepose}, gloss transcription \citep{dataset:mesch2012meaning, johnston2016, konrad2018}, sign language detection \citep{detection:borg2019sign,detection:moryossef2020real}, sign language identification \citep{identification:gebre2013automatic,identification:monteiro2016detecting}, and sign language segmentation \citep{segmentation:bull2020automatic,segmentation:farag2019learning,segmentation:santemiz2009automatic}.

\begin{figure}
    \centering
    \includegraphics[width=\linewidth]{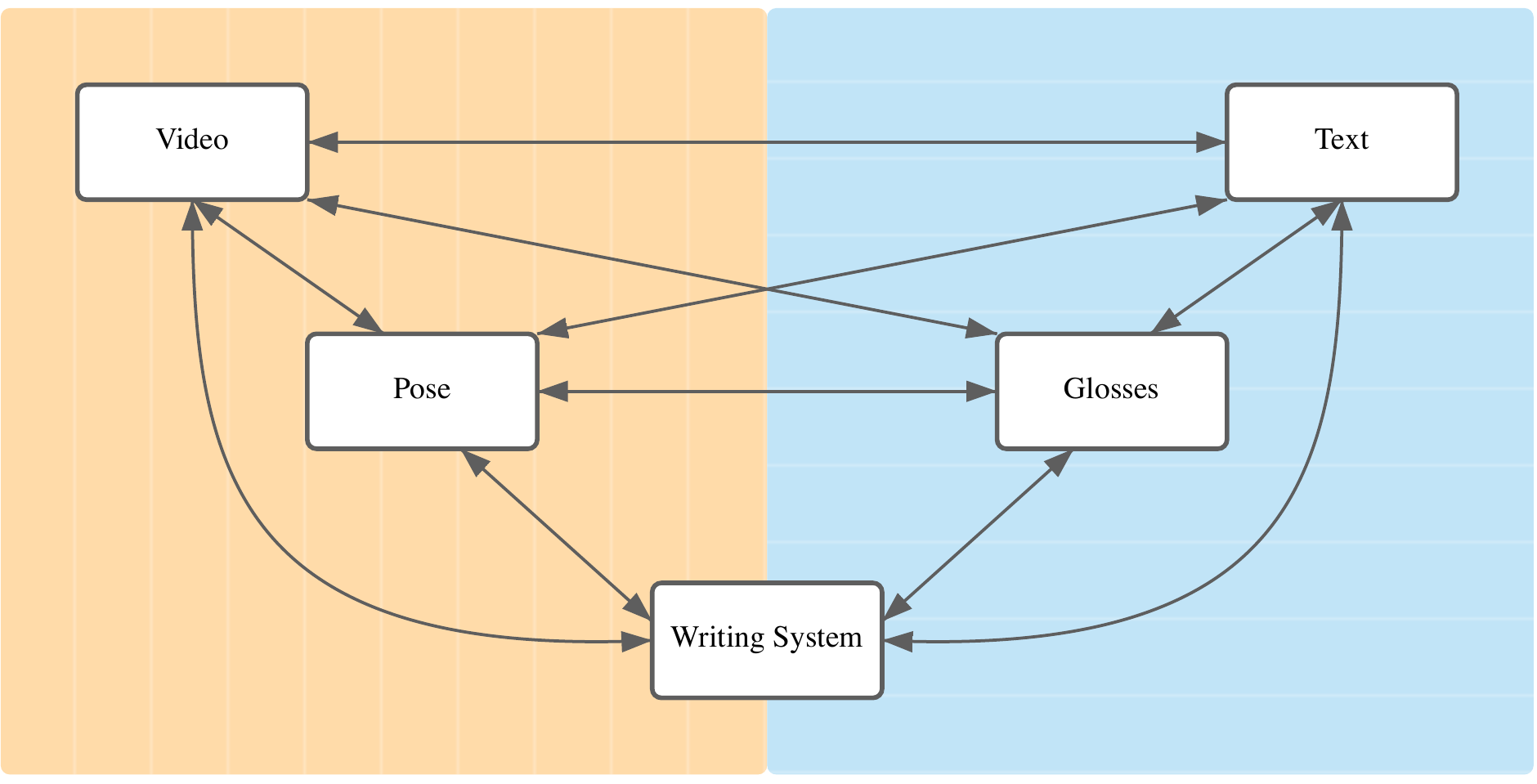}
    \caption{SLP tasks. Every edge on the left side represents a task in CV (language-agnostic). Every edge on the right side represents a task in NLP (language-specific). Every edge crossing both sides represents a task requiring a combination of CV and NLP. Figure taken from \citet{moryossef2021slp}.}
    \label{fig:slp_tasks}
\end{figure}

Besides, tasks including sign language recognition \citep{adaloglou2021comprehensive}, translation, and production involve transforming one sign language representation to another or from/to spoken language text, as shown in Figure \ref{fig:slp_tasks}\footnote{In the paper, we distinguish between a phonetic ``writing system'' (e.g., SignWriting) and ``glosses'' (lexical notation, marking the semantics of each sign with a distinct category).}. We find that existing works cover gloss-to-text \citep{cihan2018neural,yin2020better} (where ``text'' denotes spoken language text), text-to-gloss \citep{zhao2000machine,dataset:othman2012english}, video-to-text \citep{camgoz2020sign,camgoz2020multi}, pose-to-text \citep{dataset:ko2019neural}, and text-to-pose \citep{saunders2020adversarial,saunders2020everybody,saunders2020progressive,pose:zelinka2020neural,xiao2020skeleton}.

\subsection{Motivation}

Our work is the first to explore translation between spoken language text and sign language content represented in SignWriting\footnote{Related work based on HamNoSys: \citet{morrissey2011assessing, Sanaullah2021, walsh2022changing}}. We focus on a sign language writing system for the following reasons: (a) currently an end-to-end (video-to-text/text-to-video) approach is not feasible. State-of-the-art systems either have a BLEU score lower than 1 \citep{muller-etal-2022-findings} or work only on a very narrow linguistic domain, e.g., \citet{camgoz2020sign,camgoz2020multi} work on the RWTH-PHOENIX-Weather T data set which covers only 1,231 unique signs from weather reports (less than what we use in Table \ref{tab:important_puddles}); (b) a writing system is lower-dimensional than videos (not all parts of a video are relevant in a linguistic sense), while adequate to encode information of signs; (c) written sign language is a closer fit to current MT pipelines than videos or poses; (d) a phonetic writing system is a more universal solution than glossing since glosses are semantic and therefore language-specific, and are an inadequate representation of meaning \citep{muller2022considerations}.

\begin{figure*}
    \centering
    \includegraphics[width=\linewidth]{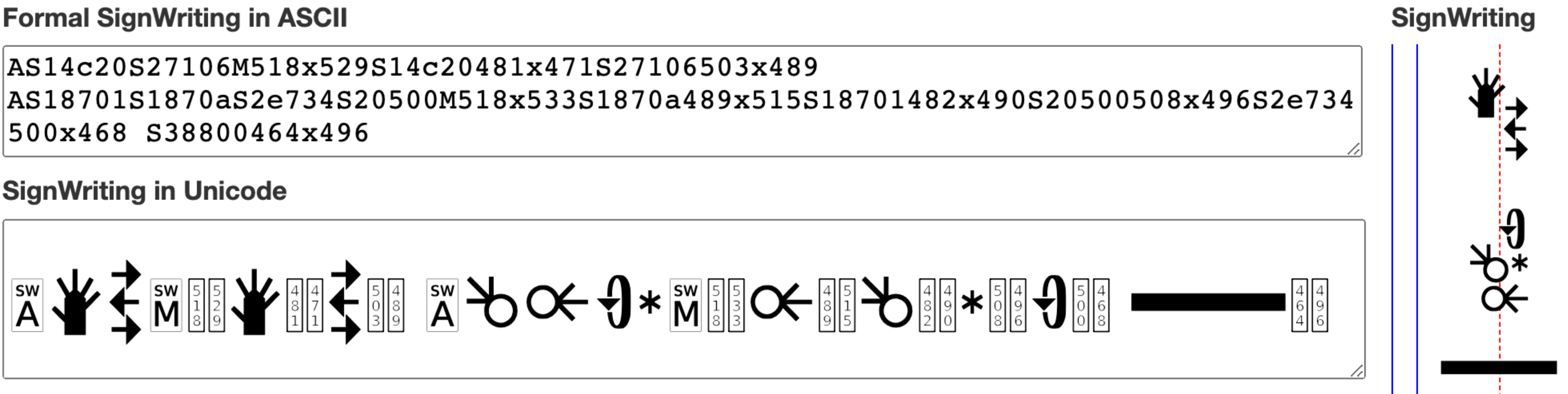}
    \caption{``Hello world.'' in FSW, SWU and SignWriting graphics. In FSW/SWU, \texttt{A}/\texttt{SWA} and \texttt{M}/\texttt{SWM} are the box markers (acting as sign boundaries); \texttt{S14c20} and \texttt{S27106} (graphemes in SWU) are the symbols; \texttt{518} and \texttt{529} are the x, y positional numbers on a 2-dimensional plane that denote symbols' position within a sign box, \texttt{S38800} (horizontal bold line in SWU) is the punctuation full stop symbol.}
    \label{fig:fsw_swu}
\end{figure*}

\subsection{SignWriting, FSW, and SWU}

SignWriting \citep{writing:sutton1990lessons} is a featural and visually iconic sign language writing system (introduced extensively in Appendix \ref{sec:SignWriting}). Previous work explored recognition \citep{Stiehl2015} and animation \citep{Bouzid2013} of SignWriting.

SignWriting has two computerized specifications, Formal SignWriting in ASCII (FSW) and SignWriting in Unicode (SWU). SignWriting is two-dimensional, but FSW and SWU are written linearly, similar to spoken languages. Figure \ref{fig:fsw_swu} gives an example of the relationship between SignWriting, FSW, and SWU\footnote{Online demonstration: \url{https://slevinski.github.io/SuttonSignWriting/characters/index.html}.}. We use FSW in our research instead of SWU to explore the potential of factorizing SignWriting symbols and utilizing numerical values of their position (\S\ref{sec:parse_fsw}, \S\ref{sec:factored}).

\section{Data and method}
\label{sec:data_and_method}

The data source we use for this research is SignBank, the largest repository of \textit{SignPuddles}\footnote{\url{https://www.signbank.org/signpuddle/}}. A SignPuddle is a community-driven dictionary where users add parallel examples of SignWriting and spoken language text (not necessarily with corresponding videos and glosses). The puddles contain material from various signed languages and linguistic domains (e.g., general literature or Bible) without a strict writing standard.
We use the Sign Language Datasets \citep{moryossef2021datasets} library to load SignBank as a Tensorflow Dataset.

\subsection{Data statistics}

In SignBank, there are roughly 220k parallel samples from 141 puddles covering 76 language pairs, yet the distribution is unbalanced (full details in Appendix \ref{sec:full_stat}). Relatively high-resource language pairs (over 10k samples) are listed in Table \ref{tab:high_resource_languages}.


\begin{table*}
\centering
\begin{tabular}{lrr}
\toprule
\textbf{language pair} & \textbf{\#samples} & \textbf{\#puddles}	\\
\midrule
en-us (American English \& American Sign Language) & 43,698 & 7 \\
pt-br (Brazilian Portuguese \& Brazilian Sign Language) & 42,454 & 3 \\
de-de (Standard German \& German Sign Language) & 24,704 & 3 \\
fr-ca (Canadian French \& Quebec Sign Language) & 11,189 & 3 \\
\bottomrule
\end{tabular}
\caption{Relatively high-resource language pair statistics.}
\label{tab:high_resource_languages}
\end{table*}

\begin{table*}
\centering
\begin{tabular}{lcrrr}
\toprule
\textbf{puddle name} & \textbf{language pair} & \textbf{\#samples} & \textbf{\#signs}& \textbf{mean sequence len}	\\
\midrule

Literature US & en-us & 700 & 9,922 & 24 \\
ASL Bible Books NLT & en-us & 11,667 & 51,485 & 24 \\
ASL Bible Books Shores Deaf Church & en-us & 4,321 & 44,612 & 31 \\
Literatura Brasil & pt-br & 1,884 & 19,221 & 13 \\
\bottomrule
\end{tabular}
\caption{Primary sentence-pair puddles. Mean sequence length is measured by the mean number of words in the spoken language sentences.}
\label{tab:important_puddles}
\end{table*}

Notably, most of the puddles are dictionaries, which we consider less valuable than sentence pairs (instances of continuous signing) for a general MT system. If dictionaries are used as training data, we expect models to memorize word mappings and not learn to generate sentences.


Therefore, we treat the four sentence-pair puddles (Table \ref{tab:important_puddles}) of the relatively high-resource language pairs as primary data and the other dictionary puddles as auxiliary data. Note that even the language pairs constituting the high-resource pairs of SignBank are low-resource compared to datasets used in mature MT systems for spoken languages, where millions of parallel sentences are commonplace \citep{akhbardeh-etal-2021-findings}.

\subsection{Data preprocessing}


We first perform general data cleaning to extract the main body of spoken language text and remove irrelevant parts such as HTML tags or samples that are empty or too long (100 words for a dictionary entry). We then learn a byte pair encoding (BPE) segmentation \citep{Sennrich2016a} on the cleaned spoken language text, using a vocabulary size of 2,000.


\paragraph{Multilingual models}
In our multilingual experiments (\S\ref{subsec:model_2}, \S\ref{subsec:model_3}), we learn a shared BPE model across all spoken languages.

Following \citet{Johnson2017}, we add special tags at the beginning of source sequences to indicate the desired target language and nature of the training data (sentence pair or dictionary).
%
Three types of tags are designed to encode all necessary information: (a) spoken language code; (b) country code\footnote{spoken language code plus country code specifies a one-to-one mapping to a related signed language in our data.}; (c) dictionary vs. sentence pair. For example, an English sentence to be translated into American Sign Language is represented as the following:

\begin{quote}\texttt{\textless2en\textgreater\ \textless4us\textgreater\ \textless sent\textgreater \, Hello world.}\end{quote}

\paragraph{Data split}

We shuffle the data and split it into 95\%, 3\%, and 2\% for training, validation, and test sets, respectively.

\subsection{FSW parsing}
\label{sec:parse_fsw}

On the sign language side, an appropriate segmentation and tokenization strategy is needed for the FSW data. We parse an original FSW sequence (e.g. Figure \ref{fig:fsw_swu}) into several pieces:

\begin{itemize}
    \item box markers: \texttt{A}, \texttt{M}, \texttt{L}, \texttt{R}, \texttt{B};
    \item symbols: \texttt{S1f010}, \texttt{S18720}, etc.;
    \item positional numbers x and y: \texttt{515}, \texttt{483}, etc.;
    \item punctuation marks (special symbols without box markers): \texttt{S38800}, etc.
\end{itemize}

We further factorize each symbol into several parts regarding its orientation (see Figure \ref{fig:orientation} in Appendix \ref{sec:SignWriting} for an explicit motivation of this step). For example, the symbol \texttt{S1f010} is split into:

\begin{itemize}
    \item symbol core: \texttt{S1f0};
    \item column number (from \texttt{0} to \texttt{5}): \texttt{1};
    \item row number (from \texttt{0} to hex \texttt{F}): \texttt{0}.
\end{itemize}

For positional numbers, which have a large range (from \texttt{250} to \texttt{750}) and are encoded discretely, we hypothesise that models might have difficulty understanding their relative order. Therefore, we further calculate two additional factors that denote a symbol's relative position (based on the absolute numbers) within a sign: relative x and relative y, both ranging from \texttt{0} to \texttt{\#symbols - 1}.


We provide a full example of the result of FSW parsing in Listing \ref{lst:factorization} in Appendix \ref{sec:full_stat}.

\subsection{Factored machine translation}
\label{sec:factored}

We use a factored machine translation system \citep{Koehn2007,GarciaMartinez2016} to encode or decode parsed FSW sequences. 
We argue that this architecture is suitable because concatenating all parsed FSW tokens results in sequences much longer than the maximum length of many Transformer models (e.g., 512).


From another perspective, the essential information units are the symbols. Nevertheless, the positional numbers are necessary to determine how symbols are assembled. The same symbols can be arranged differently in space to convey different meanings.

\begin{figure}
    \centering
    \includegraphics[width=\linewidth]{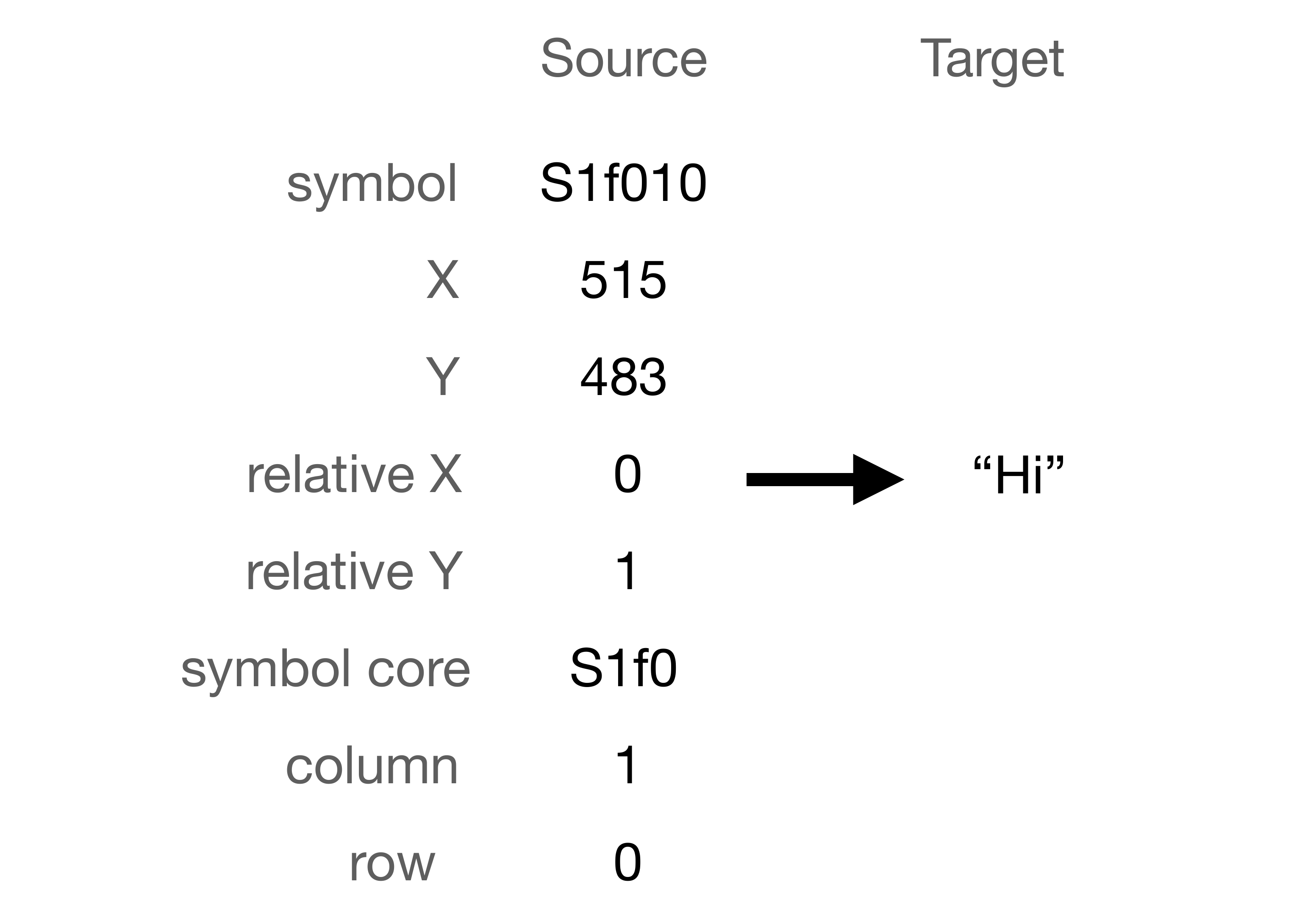}
    \caption{Representation of translating a FSW symbol together with its factors to English.}
    \label{fig:factored}
\end{figure}

In our setup, we treat the symbols (including punctuation marks and box markers) as the primary source/target tokens and the rest as source/target factors that are strictly aligned with each source/target token (illustrated by Figure \ref{fig:factored}).

Depending on the translation direction, factored FSW representations need to be encoded or decoded. For encoding
(when FSW is the source), we embed each factor separately and then concatenate them to the aligned symbol's embedding.
%
For decoding (when FSW is the target), we use only a subset of factors (absolute x and y) because others are irrelevant for prediction, and additional weighted cross-entropy losses are calculated.



\section{Experiments and results}
\label{sec:exp}

This section introduces three lines of experiments on both bilingual and multilingual SignWriting translation. We use Transformer models \citep{vaswani2017attention} that support source and target factors. See Appendix \ref{sec:exp_config} for more details on our training configuration.


\subsection{Initial exploration with a bilingual model}

\label{subsec:model_1}

For a first exploration, we train a bilingual model that translates from American Sign Language (ASL) to English (en-us). The purpose of this experiment is (a) to demonstrate that automatic SignWriting translation is feasible and (b) to explore different strategies for data processing and hyperparameters.

We use roughly 40k parallel training samples comprising roughly 15k sentence pairs and 25k dictionary pairs.
%
%
The quality of spoken language translation is measured by BLEU \citep{Papineni2002} and chrF2 \citep{Popovic2015}. Table \ref{tab:results_sign_to_en} shows the evaluation results on the test set.

\begin{table*}
\centering
\begin{tabular}{clrr}
\toprule
& \textbf{model} & \textbf{BLEU} & \textbf{chrF2}	\\
\midrule
E1 & baseline (lowercase training and test data) & 22.5 & - \\
E2 & E1 + dictionary data & 25.2 & - \\
E3 & E2 + BPE & 27.0 & 46.2 \\
E4 & E3 + x,y as factors & 27.5 & 46.5 \\
E5 & E4 + symbol core as source + row, col as factors & 23.1 & 41.2 \\
E6 & E4 + relative x, y as factors & 28.1 & 47.5 \\
E7 & E6 + aggressive dropout + tied softmax & 31.4 & 52.0 \\
E8 & E7 + symbol core, row, column as factors & 32.0 & 52.7 \\
\midrule
E9 & E8 + remove lowercasing & 30.8 & 51.2 \\
E10 & E9 + smaller BPE vocab 2000 to 1000 & 29.5 & 50.8 \\
\bottomrule
\end{tabular}
\caption{Translation quality of ASL\textrightarrow en-us bilingual models. Note that E1 to E8 are trained and evaluated with all spoken language data lowercased, while from E9 to all later experiments we remove the lowercasing, so we expect a little performance drop for the later experiments. We introduce chrF2 as an evaluation metric starting from E3.
}
\label{tab:results_sign_to_en}
\end{table*}

\subsection{Multilingual sign-to-spoken translation}
\label{subsec:model_2}

Here we extend our initial bilingual model to a multilingual setting, translating from multiple signed languages to multiple spoken languages. We define two data conditions:

\begin{itemize}
    \item \textbf{high-resource:} using roughly 100k parallel training samples (roughly 17k sentence pairs and roughly 83k dictionary pairs covering four language pairs),
    \item \textbf{adding low-resource:} in addition to the high-resource data, use all additional language pairs in SignBank that have at least 1k parallel samples (most of which are dictionaries). The total number of training examples grows to roughly 170k, covering 21 language pairs (Table \ref{tab:all_languages}).
\end{itemize}




\noindent The exact factorization strategy and model hyperparameters are informed by our bilingual experiments reported in Table \ref{tab:results_sign_to_en}.

\paragraph{Evaluating dictionary entries} For these multilingual models, many of the training samples are dictionary entries, and so are some test samples. To evaluate the translation quality for dictionary entries, we use top-n accuracy, which tests whether one of the top-n translation candidates from beam search matches the entry from the reference.


Table \ref{tab:results_sign_to_spoken} shows the evaluation results on the test set.

\begin{table*}
\centering
\begin{tabular}{llrr}
\toprule
\textbf{language} & \textbf{metrics} & \textbf{4 language pairs (100k)} & \textbf{21 language pairs (170k)}	\\
\midrule
\multirow{4}{*}{en-us (40k)} & BLEU & 29.5 & 25.0 \\
            & chrF2 & 49.8 & 47.0 \\
            & top-1 & 0.37 & 0.33 \\
            & top-5 & 0.52 & 0.45 \\
\midrule
\multirow{2}{*}{en-sg (1k)} & top-1 & - & 0.20 \\
           & top-5 & - & 0.27 \\
\midrule
\multirow{4}{*}{pt-br (40k)} & BLEU & 23.8 & 6.4 \\
            & chrF2 & 44.3 & 17.5 \\
            & top-1 & 0.12 & 0.09 \\
            & top-5 & 0.17 & 0.15 \\
\midrule
\multirow{4}{*}{mt-mt (4k)} & BLEU & - & 10.1 \\
           & chrF2 & - & 29.8 \\
           & top-1 & - & 0.05 \\
           & top-5 & - & 0.05 \\
\midrule
\multirow{2}{*}{de-de (20k)} & top-1 & 0.22 & 0.15 \\
            & top-5 & 0.31 & 0.27 \\
\midrule
\multirow{2}{*}{de-ch (4k)} & top-1 & - & 0.04 \\
           & top-5 & - & 0.06 \\
\midrule
\multirow{2}{*}{fr-ca (10k)} & top-1 & 0.04 & 0.07 \\
            & top-5 & 0.08 & 0.10 \\
\midrule
\multirow{2}{*}{fr-fr (1k)} & top-1 & - & 0.16 \\
           & top-5 & - & 0.24 \\
\midrule
\multirow{2}{*}{fr-ch (8k)} & top-1 & - & 0.07 \\
           & top-5 & - & 0.09 \\
\bottomrule
\end{tabular}
\caption{Translation quality of multilingual sign-to-spoken models (partial results on the most frequent languages). Languages without sentence pairs are only evaluated by top-n accuracy. Empty cells mean that a language pair is not supported by the model. In the parentheses are the rough numbers of samples.}
\label{tab:results_sign_to_spoken}
\end{table*}

\begin{table*}
\centering
\begin{tabular}{clrrrr}
\toprule
 & \textbf{model} & \textbf{BLEU} & \textbf{chrF2++} & \textbf{MAE x} & \textbf{MAE y} \\
\toprule
E1 & 2symbol+numbers & 6.6 & 23.1 & - & - \\
\midrule
E2 & 2symbol & 25.6 & 44.2 & - & - \\
\midrule
E3 & 2symbol+factors (w=1) & 19.9 & 39.1 & 46.5 & 52.6 \\
E4 & 2symbol+factors (w=0.5) & 21.9 & 40.8 & 46.8 & 52.7 \\
E5 & 2symbol+factors (w=0.2) & 22.9 & 42.0 & 47.4 & 53.0 \\
E6 & 2symbol+factors (w=0.1) & 22.0 & 41.7 & 46.4 & 52.2 \\
E7 & 2symbol+factors (w=0.01) & 21.0 & 40.9 & 48.4 & 58.3 \\
\bottomrule
\end{tabular}
\caption{Translation quality of multilingual spoken-to-sign models. Evaluated in BLEU (on symbol, higher is better), chrF2++ (on symbol, higher is better), and MAE (on positional numbers, lower is better). \texttt{w} denotes the weight between each factor's loss and the main target loss.}
\label{tab:results_spoken_to_sign}
\end{table*}

\begin{table*}
\centering
\begin{tabular}{lrr}
\toprule
\textbf{language} & \textbf{BLEU (on symbols)} & \textbf{chrF2++ (on symbols)} \\
\midrule
en-us (40k) & 35.7 & 58.4 \\
pt-br (40k) & 1.9 & 14.9 \\
de-de (20k) & 17.3 & 43.2 \\
fr-ca (10k) & 5.3 & 19.1 \\
\bottomrule
\end{tabular}
\caption{Translation quality of multilingual spoken-to-sign model (w=0.1) per language. In the parentheses are the rough numbers of samples per language.}
\label{tab:results_spoken_to_sign_multilingual}
\end{table*}

\subsection{Multilingual spoken-to-sign translation}
\label{subsec:model_3}

Finally, we train multilingual models that translate in the reverse direction, from spoken languages to signed languages. The data and model configuration are the same as for the multilingual sign-to-spoken model under high-resource data condition.

\paragraph{FSW decoding strategies} SignWriting utterances are parsed into a factored FSW representation (\S\ref{sec:parse_fsw}, \S\ref{sec:factored}) and are used for encoding successfully, yet it is not obvious how to best decode to FSW.
We try the following strategies: (a) predicting everything (including positional numbers) as target tokens all in one long target sequence, inspired by \citet{Chen2022}; (b) predicting symbols only (as a comparative experiment); (c) predicting symbols with positional numbers as target factors.

During decoding in the test phase, we apply beam search only for the main target token prediction. Target factors do not participate in beam search, i.e., each target factor prediction is the argmax of the corresponding output layer distribution. We shift target factors to the right by 1 to condition their prediction on the previously generated target symbol.

\paragraph{Evaluation of FSW output}

Due to variations of SignWriting symbols based on different orientations that do not change meaning (Figure \ref{fig:orientation}), evaluating FSW output only at the token (symbol) level is not sufficient. Therefore, we evaluate the output symbols (e.g., line 4 in Listing \ref{lst:factorization}) not only with BLEU, but also chrF2++, which captures both word-level and character-level statistics.
%
%
Additionally, we evaluate the output positional numbers by mean absolute error (MAE) to measure the distance between predicted positional numbers and the ones from the FSW reference (e.g., lines 5 and 6 in Listing \ref{lst:factorization}). Let $x$ be the predicted sequence of positional numbers and $y$ be the gold sequence:

\[ MAE(x, y) = \frac{1}{|x|} \sum_{i=1}^{|x|}|x_i-y_i| \]

\noindent where if the predicted and gold sequences do not have the same length, they are padded with zeros.


Table \ref{tab:results_spoken_to_sign} shows the results of evaluation on the test set. Table \ref{tab:results_spoken_to_sign_multilingual} shows the results of multilingual evaluation on \emph{E6} (w=0.1) of Table \ref{tab:results_spoken_to_sign}.

\section{Discussion}
\label{sec:discussion}

\subsection{Effect of adding dictionaries, BPE, and low-resource optimizations}


As shown in Table \ref{tab:results_sign_to_en},  enlarging the sentence-level training data (15k sentence pairs) with 25k dictionary pairs improves the translation quality by 2.7 BLEU (\emph{E1} vs. \emph{E2}). Likewise, applying BPE segmentation to the spoken language side also improves translations by 1.8 BLEU (\emph{E2} vs. \emph{E3}).

We evaluate several low-resource ``tricks'' \citep{Sennrich2019} including aggressive dropout and weight tying\footnote{
The tying is only between the target embedding and the softmax output matrix since the source and target languages are of a very different nature and therefore cannot be tied.}. These low-resource optimizations borrowed from spoken language MT prove to be effective for sign language translation as well, as they result in an improvement of 3.3 BLEU and 4.5 chrF2 (\emph{E6} vs. \emph{E7} in Table \ref{tab:results_sign_to_en}).

\subsection{Utilizing positional numbers}
\label{subsec:utilize_numbers}

In earlier sections we introduce novel methods to parse and factorize FSW (\S\ref{sec:parse_fsw}, \S\ref{sec:factored}). However,  from a model training perspective it is unclear how to best utilize additional factors such as positional numbers. In \emph{E4}, \emph{E5}, \emph{E6}, and \emph{E8} of Table \ref{tab:results_sign_to_en}, we explore different ways of including factors.

We find that the best strategy is explicitly adding all additional information (x, y, relative x, relative y, symbol core, column number, row number) as source factors while keeping symbols as the primary source tokens. This strategy achieves the state-of-the-art performance of 32.0 BLEU and 52.7 chrF2 in \emph{E8}.

\subsection{Generating positional numbers}
\label{subsec:generate_numnber}

We explore different ways of generating positional numbers in Table \ref{tab:results_spoken_to_sign}. As a first attempt, we treat them as normal target tokens in \emph{E1}, which results in poor performance and overly long target sequences, and long beam search decoding time.

In all subsequent experiments, we treat positional numbers as target factors and generally achieve over 20 BLEU (evaluating on symbols). In \emph{E2}, we translate only the symbols as a baseline. Then we also try translating with target factors and varying the weights between factors and the primary target, i.e., symbols. Finally, we observe that \emph{E6} (w=0.1) leads to the best trade-off between generating symbols and positional numbers.

\subsection{Multilingual performance}

We discuss multilingual performance mainly based on Table \ref{tab:results_sign_to_spoken}. Generally speaking, the more resources a language has in the multilingual model, the better its performance \citep{zhou2021distributionally}. The two target languages most frequent in the training data--American English (en-us, 40k) and Brazilian Portuguese (pt-br, 40k)--have the highest translation quality.

\paragraph{Multilingual transfer effects}

We observe examples of both positive and negative multilingual transfer.
Evidence shows that a relatively high-resource language can help a related low-resource language. For instance, the performance of Singaporean English (en-sg, 1k) is likely improved by American English (en-us, 40k), which is almost as good as Standard German (de-de, 20k).


The comparable bilingual en-us model (\emph{E9} in Table \ref{tab:results_sign_to_en}) outperforms en-us in our multilingual sign-to-spoken model in Table \ref{tab:results_sign_to_spoken} by 1.3 BLEU and 1.4 chrF2.
However, when extending the training data from 4 to 21 language pairs, we observe severe degradations: en-us drops by 4.5 BLEU and 2.8 chrF2; Brazilian Portuguese (pt-br) drops by 17.4 BLEU and 26.8 chrF2.

Such findings are in line with previous work on highly multilingual translation systems. For example, \citet{Aharoni2019} finds that average per-language performance drops when the number of languages increases. We conclude that SignWriting translation suffers from a similar \textit{curse of multilinguality}.


\subsection{Side-by-side SignWriting example}
\label{subsec:side_by_side}

\begin{figure}
    \centering
    \includegraphics[width=\linewidth]{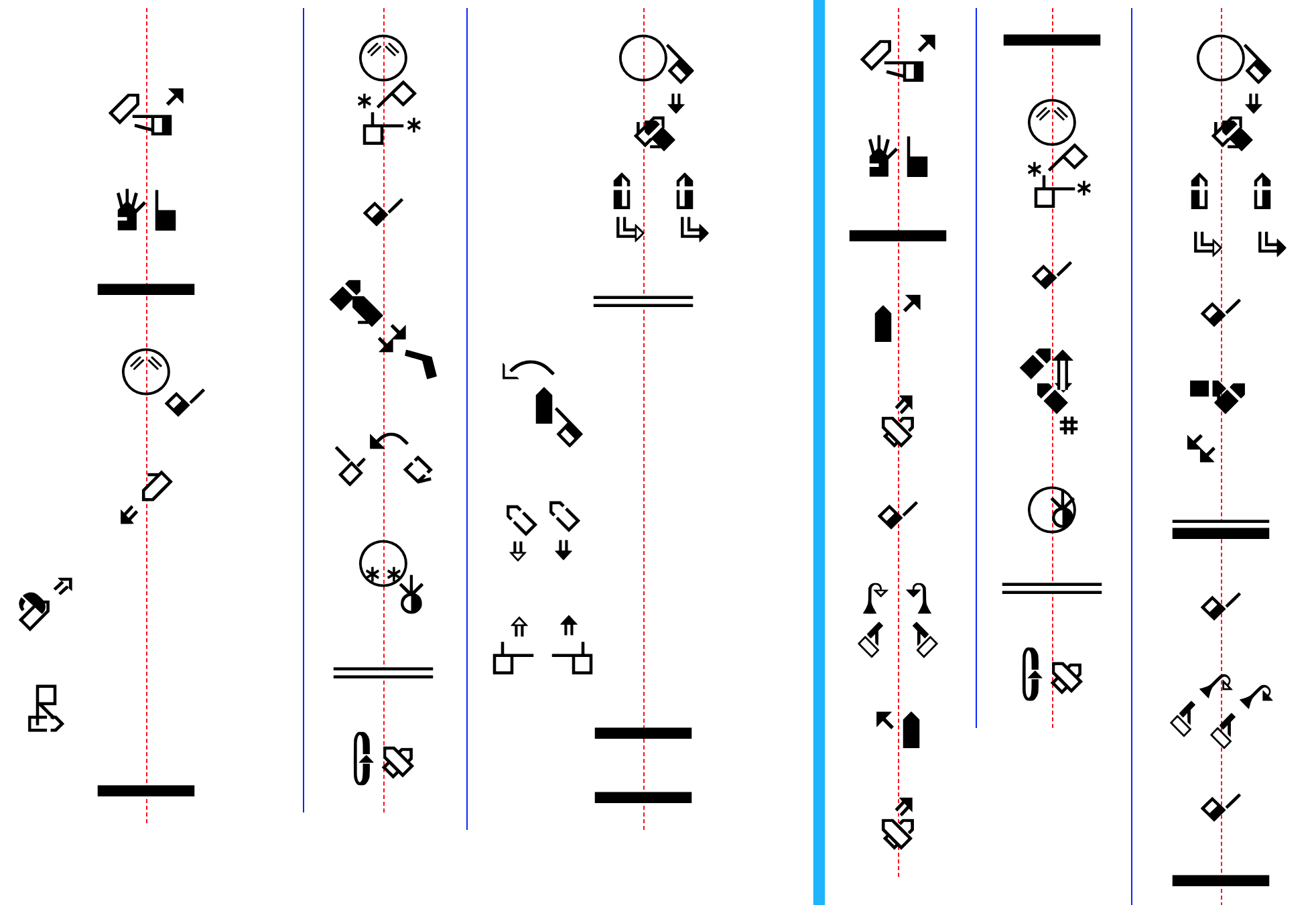}
    \caption{Side-by-side SignWriting example. ASL translation of the English sentence ``Verse 41. He gave her his hand and helped her up. Then he called in the widows and all the believers, and he presented her to them alive.'' Separated by the vertical bold (light blue) line, the left is the gold sentence, and the predicted sentence is on the right. The predicted sentence translated back to English is ``Verse 41. His hand he gives her hand. Then he helped up, all believers he warned: he put there.''}
    \label{fig:side_by_side}
\end{figure}

Finally, to gain intuition for how well the translation model signs, we give a side-by-side example of SignWriting graphics. We compare the reference and model prediction of an ASL utterance corresponding to an American English utterance from the Bible corpus, shown in Figure \ref{fig:side_by_side}.

We ask an ASL user proficient in SignWriting for a translation of the predicted SignWriting back to English to assess the quality of the prediction.

Similar patterns appear in both: in the beginning, the model signs ``Verse 41'' in the same way as in the reference; the graphics in the top parts of all the columns are consistent; and we see correct symbols sometimes predicted with slightly different positions.

More translation examples can be seen in Appendix \ref{sec:side_by_side_more}.

\section{Conclusion}

This work explores building bilingual and multilingual translation systems between spoken and signed languages. Instead of representing sign language as videos (or as continuous features derived from videos) common in previous research, we propose to represent sign language in SignWriting, a sign language writing system. We argue that using a written form is more amenable to well-established NLP techniques.

However, encoding or decoding SignWriting in an MT system requires specialized tools. Therefore, we propose novel methods to parse, factorize, decode, and evaluate SignWriting sequences. Our factorization technique divides SignWriting sequences into meaningful units such as sign symbols and positional numbers. The factors are then encoded or decoded by a factored Transformer model.

As a result, we achieve over 30 BLEU in the bilingual setting and over 20 BLEU for some high-resource language pairs in both directions in the multilingual setting.

Using SignWriting as an intermediate representation enables us to reuse tools (e.g., evaluation metrics) from spoken language translation. We also observe striking similarities to spoken language MT in the experiments themselves. For example, low-resource optimizations have a similar impact, and multilingual models exhibit similar transfer effects. These findings validate our use of an intermediate text representation for signed languages to include them in NLP research.

\section{Limitations}

\subsection{A word on top-n accuracy}

In the translation of dictionary data, if a dictionary entry has been seen during training, assuming the model has enough capacity, it should memorize and predict it. However, evaluating the translation is tricky, so in \S\ref{subsec:model_2} we resort to using top-n accuracy.

Paradoxically, high top-n accuracy on the test set does not guarantee good generalization and might be associated with overlap between the training and test sets.
Conversely, claiming a model is terrible when it performs poorly on the test set is unjustified, as there might be no overlap between the training and test sets. 
If a model has seen all the words from a language, it should perform well on whatever dictionary test set. However, this is not the case in our low-resource setting.


\subsection{Fingerspelling tokenization}
\label{subsec:future_fngr}

Fingerspelling \citep{battison1978lexical,wilcox1992phonetics,brentari2001language} is an interesting linguistic phenomenon where a signed language geographically coexists with a spoken language. 
For words with no associated signs (e.g., names of people, locations, organizations, etc.), sign language users borrow a word of a spoken language by spelling it letter-by-letter with predefined signs for the letters of the alphabet of that language. The fingerspelling (manual) alphabet of a sign language draws on a closed set of hand shapes, which are supported by SignWriting.

As fingerspelling is usually applied on a character level (rarely extending to the level of multiple characters, such as ``CH'' or ``SCH'' for the finger alphabet of Swiss German Sign Language), the way BPE segmentation works (on subword level) does not apply perfectly. However, if we could detect fingerspelling during the segmentation/tokenization process, then force fingerspelled words to be split letter-by-letter, our models should be able to learn better the mappings between fingerspelling signs and spoken language letters.

\subsection{Towards better multilingual models}

As shown by Table \ref{tab:important_puddles}, the data we use to train our models only contains many sentence pairs for American English and American Sign Language. For other language pairs, we train mainly on dictionary data. 

At the time of writing, we find a multilingual parallel corpus created from translations of the Bible\footnote{\url{https://github.com/christos-c/bible-corpus}} \citep{christ2014}, which, if aligned correctly, can be used to translate the $\sim$15k American Sign Language biblical text to another 100+ spoken languages. We believe we could train better multilingual translation models (at least on the spoken language side) based on them.

\subsection{Regression objective for positional numbers}

In our experiments, positional numbers are treated as target factors (\S\ref{subsec:generate_numnber}), contributing cross-entropy loss to the training process. However, we are aware that the positional numbers are, by nature, numeric values, so a regression objective/loss would possibly work better than the current cross-entropy loss, as it better reflects the numeric relationship between positional numeric values.

As for now, the target factor function we use is only implemented with a classification objective (cross-entropy loss). We envision that custom implementation of the regression objective might improve translation quality in this scenario.

\subsection{Possibly flawed positional number evaluation}
\label{subsec:number_evaluation}

We note that using MAE for evaluating positional numbers (\S\ref{subsec:model_3}) is possibly flawed because the predicted symbol sequences can deviate from the gold symbol sequences. If this is the case, making a token-by-token comparison on the positional numbers is meaningless, as even the sequence length can mismatch.

\subsection{Advanced SignWriting evaluation}

Finally, we call for advanced and novel methods of SignWriting evaluation, considering its differences from spoken languages.

In our experiments, we separate the evaluation of FSW symbols and positional numbers. For symbols, we borrow BLEU and chrF2/chrF2++ from spoken language evaluation since FSW symbols are the basic graphemes in SignWriting that show many similar linguistic features as spoken language words. For positional numbers, MAE is used, and its limitation is discussed in \S\ref{subsec:number_evaluation}.


From a broader perspective, FSW is merely a linearized specification of SignWriting, which means we can also evaluate on the original graphical form, as we do manually in \S\ref{subsec:side_by_side}. Moreover, we can exploit CV techniques to do an automatic comparative evaluation between predicted SignWriting graphics and gold SignWriting graphics.

Ideally, a cascading evaluation method is applied to SignWriting: we first evaluate the overall graphics of the signs, then the symbols within the signs, then the position of the symbols, then the factorized representation of the symbols. Finally, a thorough human evaluation is needed to gain better insight.

\section*{Note on reproducibility}

We will release the source code and documentation to train our models, an API server with the trained models, and a demo Web application. This will allow others to see and consistently reproduce our results with minimal changes. We encourage the community to attempt to reproduce our results and publish the results.


\section*{Acknowledgements}

This work is funded by the following projects: EASIER (Grant agreement number 101016982) and IICT (Grant agreement number PFFS-21-47). We are grateful for their support. We also thank Rico Sennrich for his suggestions.

\bibliography{custom}
\bibliographystyle{acl_natbib}

\clearpage
\appendix

\section{Extended introduction to SignWriting}
\label{sec:SignWriting}

\begin{figure}
    \centering
    \includegraphics[width=0.75\linewidth]{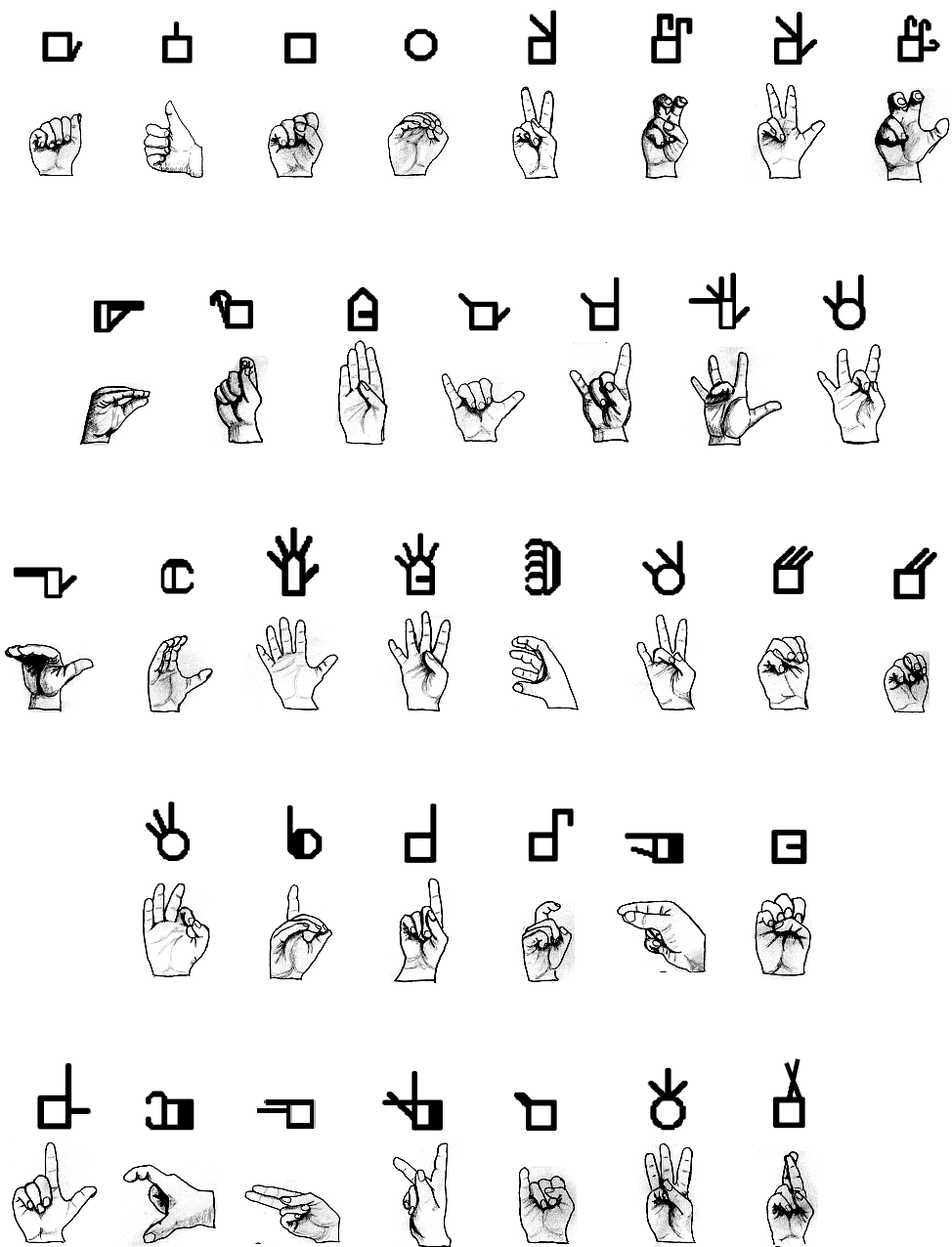}
    \caption{Hand shapes and their equivalents in SignWriting.}
    \label{fig:handshapes}
\end{figure}

\begin{figure}
    \centering
    \includegraphics[width=0.9\linewidth]{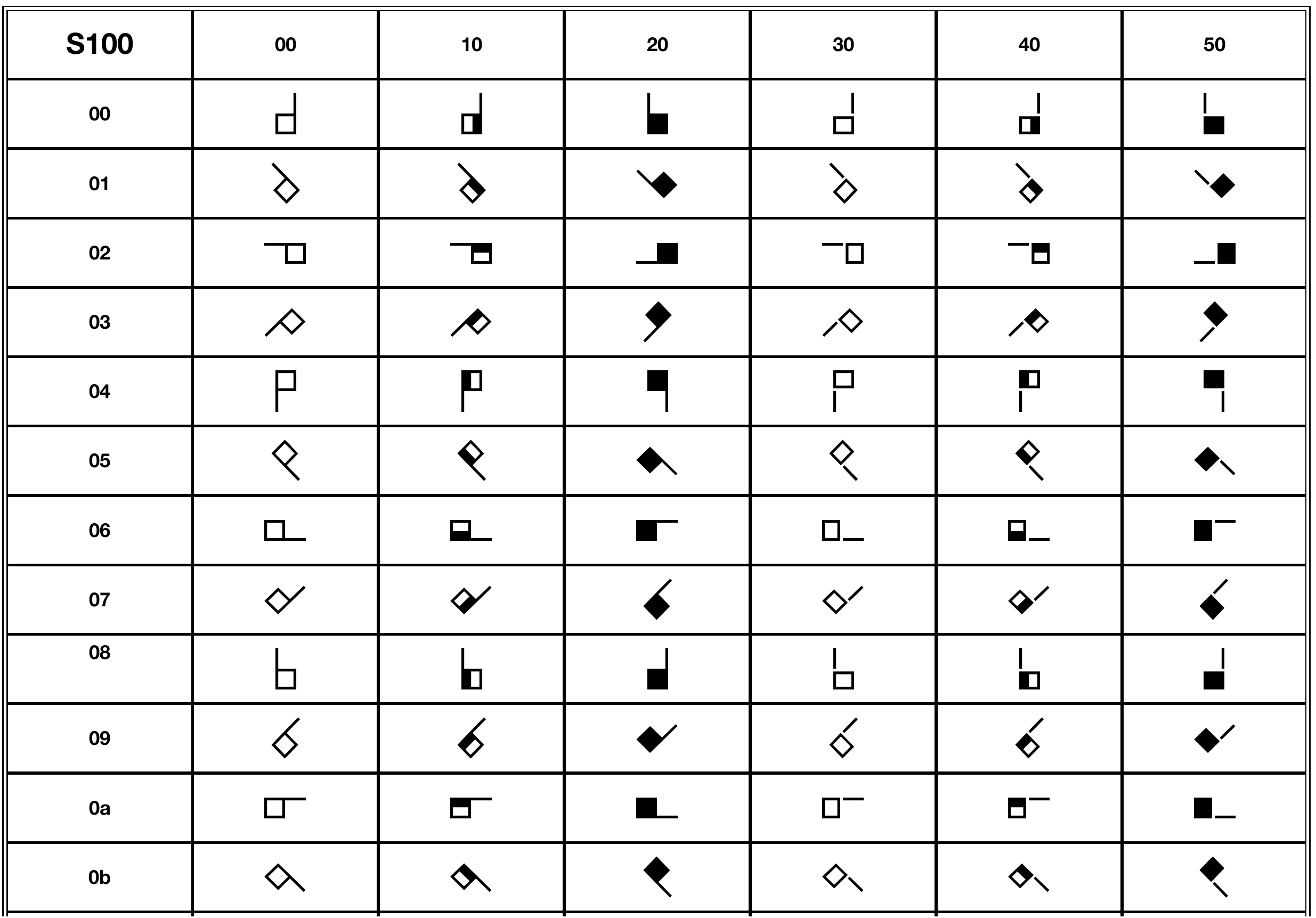}
    \caption{Orientation of a symbol in SignWriting in 3D space. Each row applies a rotation of the palm in a 2D space \textbf{vertical} to the ground. Each column applies a rotation of the palm in a 2D space \textbf{parallel} to the ground. This can be seen as a factorization of the symbol \texttt{S100xx} to its core \texttt{S100} plus row and column numbers.}
    \label{fig:orientation}
\end{figure}

\begin{figure}
    \centering
    \includegraphics[width=0.9\linewidth]{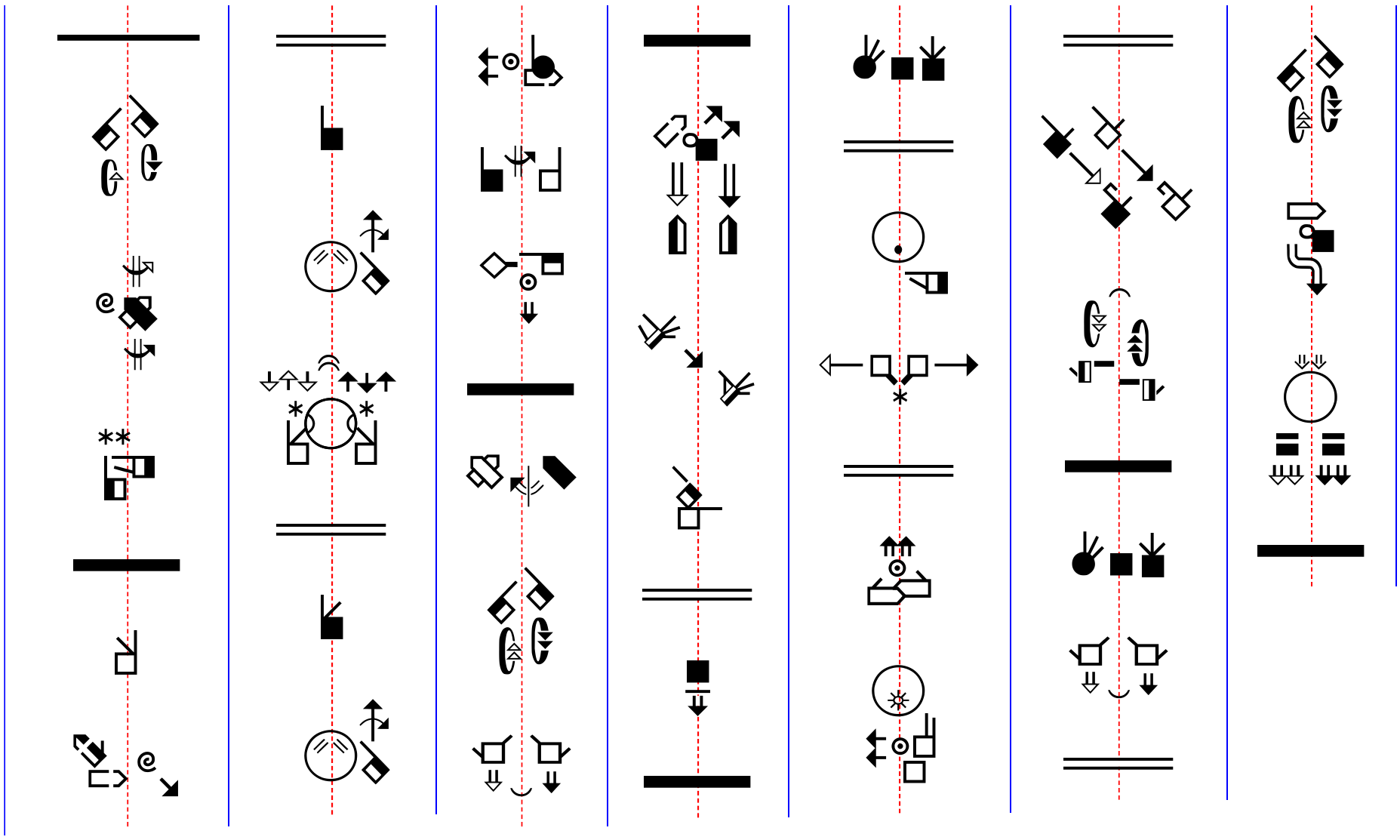}
    \caption{An example of SignWriting written in columns, ASL translation of an introduction to Formal SignWriting in ASCII. The relative positions of the symbols within the box iconically represent the locations of the hands and other parts of the body involved in the sign being represented.}
    \label{fig:arrangement}
\end{figure}

SignWriting \citep{writing:sutton1990lessons} is a sign language writing system developed by Valerie Sutton\footnote{Valerie Sutton: \url{https://en.wikipedia.org/wiki/Valerie_Sutton}} and currently managed by Steve Slevinski\footnote{Steve Slevinski: \url{https://steveslevinski.me/}}. SignWriting is very featural and visually iconic, both in:

\begin{itemize}
    \item the shapes of the symbols, which are abstract pictures of hand shapes (Figure \ref{fig:handshapes}), orientation (Figure \ref{fig:orientation}), body locations, facial expressions, contacts, and movement;
    \item the symbols' two-dimensional spatial arrangement in an invisible ``sign box'' (Figure \ref{fig:arrangement}).
\end{itemize}

Outside each sign, the script is written linearly to reflect the temporal order of signs. Signs are mostly written vertically, arranged from top to bottom within each column, interspersed with special punctuation symbols (horizontal lines), and the columns progress left to right across the page. Within each column, signs may be vertically aligned to the center or shifted left or right to indicate body shifts.

\subsection{Formal SignWriting in ASCII (FSW)}

In 2012, Formal SignWriting in ASCII (FSW) specification \citep{slevinski-formal-signwriting-08} was released and documented in an Internet Draft submitted to the IETF.


The design of FSW is computerized so that it can be recognized and processed by programs. While signed languages are natural languages, FSW is a formal language handy in mathematics, computer science, and linguistics.

Although SignWriting is two-dimensional, FSW is written linearly like spoken languages. Each sign is written as first a box marker, then a sequence of symbols, and their relative position, as illustrated by Figure \ref{fig:fsw_swu}.

\subsection{SignWriting in Unicode (SWU)}

In 2017, SignWriting in Unicode (SWU) specification \citep{slevinski-formal-signwriting-08} was released, making SignWriting included in the Unicode Standard. The Unicode block for SWU is \texttt{U+1D800} - \texttt{U+1DAAF}.

As illustrated in Figure \ref{fig:fsw_swu}, SWU is also written linearly. FSW and SWU are isomorphic and interchangeable, and both faithfully encode the complete information of SignWriting.

\newpage

\section{Experimental setup}
\label{sec:exp_config}

We performed all our experiments with Python 3.8.11 on an Nvidia Tesla V100 GPU (32GB GPU ram).

\subsection{40k sign-to-en-us}

Data includes:
\begin{itemize}
    \item $\sim$15k sentence pairs from 3 en-us puddles: Literature US, ASL Bible Books NLT, ASL Bible Books Shores Deaf Church,
    \item $\sim$25k dictionary pairs from 3 en-us puddles: Dictionary US, LLCN \& SignTyp, ASL Bible Dictionary,
\end{itemize}

\noindent which leads to:

\begin{itemize}
    \item $\sim$6k source vocabulary size (number of non-factorized symbols),
    \item $\sim$2k target vocabulary size (determined by BPE).
\end{itemize}

\noindent Final model configuration:

\begin{itemize}
    \item 6 layers + 8 heads + 512 embedding size (16 for each factor) (0.5 dropout) + 512 hidden size (0.5 dropout) + 2,048 feed forward size (0.5 dropout),
    \item initial learning rate 0.0001, decrease learning rate by a factor of 0.7 every 5 times validation score (BLEU) not improved,
    \item batch size 32 sentences, label smoothing 0.2, epochs 300,
    \item for testing, decoding with a checkpoint with the best validation score, beam size 5, alpha for length penalty 1.
\end{itemize}

Experiments were conducted with a custom version of Joey NMT \citep{Kreutzer2019} to support source factors. Each model ($\sim$47 million parameters) finished training within 1 day. 

\subsection{100k sign-to-spoken}

Data includes:
\begin{itemize}
    \item $\sim$17k sentence pairs from 3 en-us puddles (Literature US, ASL Bible Books NLT, ASL Bible Books Shores Deaf Church) and 1 pt-br puddle (Literatura Brasil),
    \item $\sim$83k dictionary pairs from 3 en-us puddles (Dictionary US, LLCN \& SignTyp, ASL Bible Dictionary), 2 pt-br puddles (Dicionário Brasil, Enciclopédia Brasil), 1 de-de puddle (Wörterbuch DE) and 1 fr-ca puddle (Dictionnaire Quebec),
\end{itemize}

\noindent which leads to:
\begin{itemize}
    \item $\sim$11k source vocabulary size (number of non-factorized symbols),
    \item $\sim$2k target vocabulary size (determined by BPE).
\end{itemize}

\noindent A little change to the previous configuration to make training more efficient:

\begin{itemize}
    \item batch size 4,096 tokens.
\end{itemize}

Experiments were conducted with a custom version of Joey NMT to support source factors. Each model ($\sim$50 million parameters) finished training within $\sim $1.5 days and $\sim$3 days, respectively.

\subsection{100k spoken-to-sign}

Data and model configurations are the same as \textbf{100k sign-to-spoken}, except that we use perplexity \citep{chen-goodman-1996-empirical,Sennrich2012} as validation score instead of BLEU.

Experiments were conducted with Sockeye \citep{hieber-etal-2020-sockeye} for the convenience of ready-to-use target factor support. Each model ($\sim$60 million parameters) finished training within $\sim$0.5 day.

\onecolumn

\section{Data}
\label{sec:full_stat}

\begin{figure*}[!htbp]
    \centering
    \includegraphics[width=\textwidth]{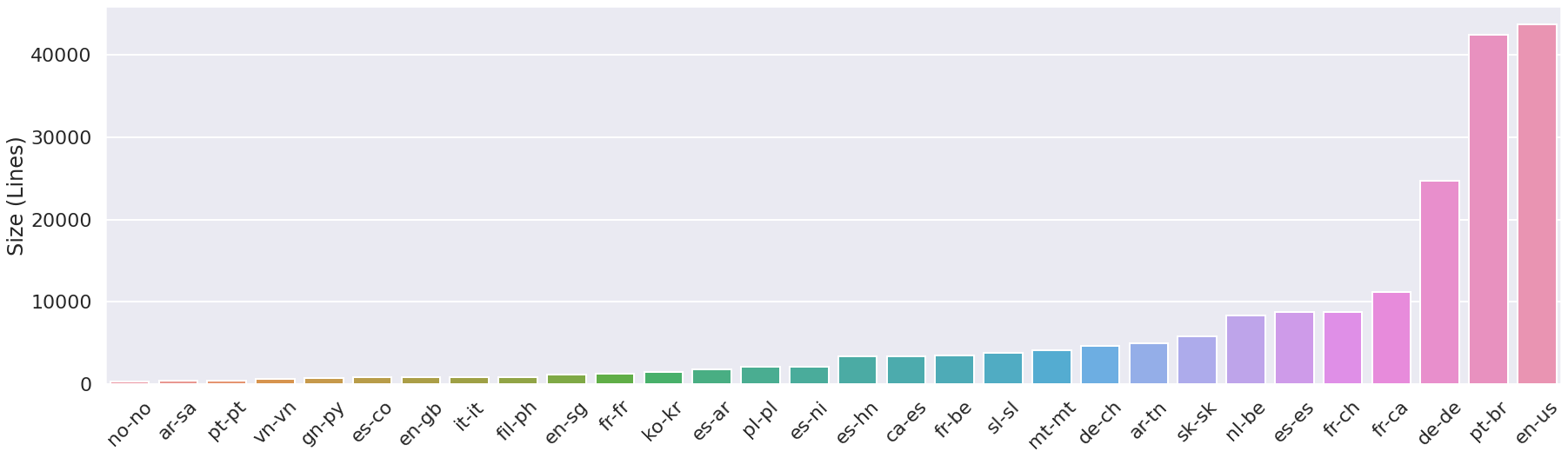}
    \caption{Data distribution (the first 30 language pairs).}
    \label{fig:data_distribution}
\end{figure*}

Figure \ref{fig:data_distribution} visualizes the language pair distribution in SignBank. Table \ref{tab:all_languages} contains an exhaustive list of all 21 language pairs used in this research. Listing \ref{lst:factorization} shows an example of FSW parsing and factorization.

\begin{table*}[!htbp]
\centering
\begin{tabular}{lrrc}
\toprule
\textbf{language} & \textbf{\#samples} & \textbf{\#puddles} & \textbf{sentence pairs (\textgreater 1k)}	\\
\midrule
en-us (American English) & 43,698 & 7 & \checkmark \\
en-sg (Singaporean English) & 1,136 & 2 \\
\midrule
pt-br (Brazilian Portuguese) & 42,454 & 3 & \checkmark \\
\midrule
mt-mt (Maltese Maltese) & 4,118 & 4 & \checkmark \\
\midrule
de-de (German German) & 24,704 & 3 \\
de-ch (Swiss German) & 4,700 & 2 \\
\midrule
fr-ca (Canadian French) & 11,189 & 3 \\
fr-ch (Swiss French) & 8,806 & 3 \\
fr-be (Belgian French) & 3,439 & 1 \\
fr-fr (French French) & 1,299 & 2 \\
\midrule
es-es (Spanish Spanish) & 8,806 & 2 \\
es-hn (Honduran Spanish) & 3,399 & 1 \\
es-ni (Nicaraguan Spanish) & 2,150 & 2 \\
es-ar (Argentinian Spanish) & 1,774 & 2 \\
\midrule
ar-tn (Tunisien Arabic) & 4,965 & 2 \\
\midrule
ca-es (Spanish Catalan) & 3,419 & 2 \\
\midrule
ko-kr (Korean Korean) & 1,525 & 1 \\
\midrule
nl-be (Belgian Flemish) & 8,301 & 2 \\
\midrule
pl-pl (Polish Polish) & 2,130 & 2 \\
\midrule
sk-sk (Czech Czech) & 5,780 & 2 \\
\midrule
sl-sl (Slovenian Slovenian) & 3,808 & 2 \\
\bottomrule
\end{tabular}
\caption{All 21 language pairs (spoken languages with corresponding signed languages).}
\label{tab:all_languages}
\end{table*}

\newpage
\begin{lstlisting}[xleftmargin=0.4cm, language=Python, caption=An example of FSW parsing and factorization., label={lst:factorization}]
{
 'fsw': 'M550x535S32a00482x483S15d09455x499S15d01522x497S22114516x484
    S22114456x484S20f00524x522S20f00451x523',
 'symbol': 'M S32a00 S15d09 S15d01 S22114 S22114 S20f00 S20f00',
 'feat_x': '550 482 455 522 516 456 524 451',
 'feat_y': '535 483 499 497 484 484 522 523',
 'feat_x_rel': '-1 3 1 5 4 2 6 0',
 'feat_y_rel': '-1 0 4 3 1 2 5 6',
 'feat_core': 'M S32a S15d S15d S221 S221 S20f S20f',
 'feat_col': '-1 0 0 0 1 1 0 0',
 'feat_row': '-1 0 9 1 4 4 0 0',
}
\end{lstlisting}

\section{More side-by-side SignWriting examples}
\label{sec:side_by_side_more}

Separated by the vertical bold (light blue) line, the left is the gold sentence, and the predicted sentence is on the right.

\begin{figure*}[!htbp]
    \centering
    \includegraphics[width=0.75\linewidth]{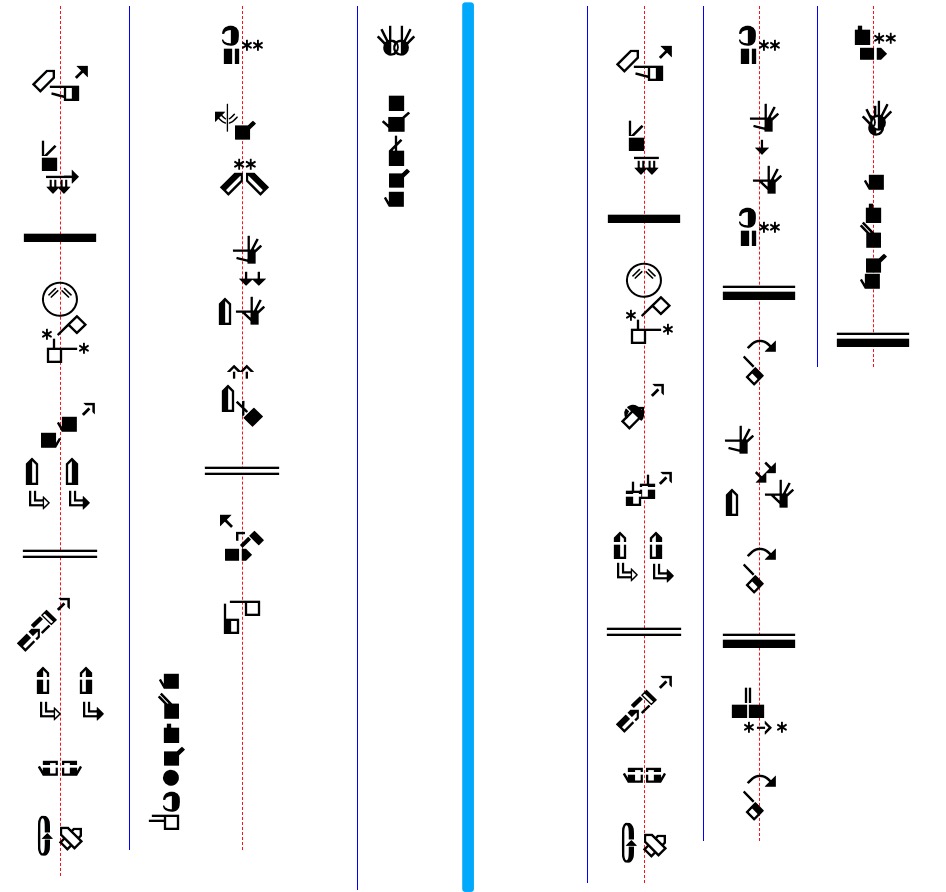}
    \caption{ASL translation of the English sentence ``Verse 22. Then the apostles and elders together with the whole church in Jerusalem chose delegates, and they sent them to Antioch of Syria''}
    \label{fig:side_by_side_2}
\end{figure*}

\begin{figure*}[!htbp]
    \centering
    \includegraphics[width=0.7\linewidth]{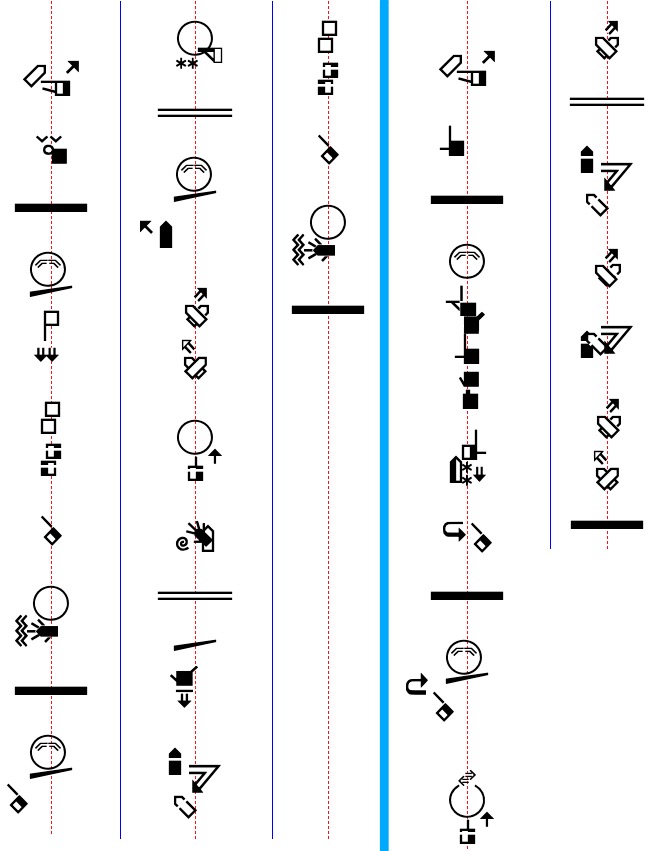}
    \caption{ASL translation of the English sentence ``These are what defile you. Eating with unwashed hands will never defile you.''}
    \label{fig:side_by_side_3}
\end{figure*}

\begin{figure*}[!htbp]
    \centering
    \includegraphics[width=0.5\linewidth]{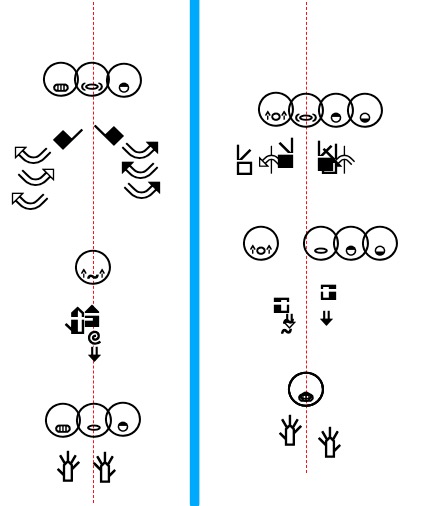}
    \caption{German Sign Language translation of the German words ``signen, senken, zehn''}
    \label{fig:side_by_side_4}
\end{figure*}

\end{document}